\documentclass[11pt]{article}

\usepackage[T1]{fontenc}
\usepackage[utf8]{inputenc}
\usepackage{lmodern}
\usepackage{microtype}
\usepackage{geometry}
\usepackage{amsmath,amssymb,amsthm}
\usepackage{enumitem}
\usepackage{natbib}
\usepackage[hidelinks]{hyperref}
\usepackage{url}

\hypersetup{
  pdftitle={Verification abundance, adjudication scarcity: what happens to mathematical knowledge when proof checking becomes free},
  pdfauthor={Maher Kallel and Mohamed El Louadi},
  pdfsubject={Formal verification, autoformalization, and mathematical knowledge},
  pdfkeywords={formal verification, autoformalization, philosophy of mathematical practice, AI generated proof, proof assistants, epistemology of testimony}
}

\begin{document}

\title{\textbf{Verification abundance, adjudication scarcity: what happens to mathematical knowledge when proof checking becomes free}}

\author{
Maher Kallel\\
Senior Strategic Consultant\\
Email: \href{mailto:maher.kallel@gmail.com}{maher.kallel@gmail.com}\\
ORCID: \href{https://orcid.org/0009-0006-2615-4392}{0009-0006-2615-4392}
\and
Mohamed El Louadi\\
Institut Supérieur de Gestion, Université de Tunis\\
41 rue de la Liberté, Cité Bouchoucha, 2000 Le Bardo, Tunisia\\
Email: \href{mailto:mohamed.louadi@isg.rnu.tn}{mohamed.louadi@isg.rnu.tn}\\
ORCID: \href{https://orcid.org/0000-0003-1321-4967}{0000-0003-1321-4967}
}

\date{}

\maketitle


\begin{abstract}
In May 2026 an OpenAI model produced a counterexample to the Erdős unit distance conjecture. Five mathematicians published a digested, human verified version of the argument the same day, and the result was absorbed into the literature within weeks. In August 2026 the same laboratory published ten results in mathematics and theoretical computer science, each accompanied by a machine checkable Lean 4 certificate with no unproved steps. Four weeks later, one of those results is the subject of an unresolved dispute in which the mathematical community has been unable to determine whether the formalization means what it claims to mean.

We argue that the difference between these outcomes is structural rather than accidental, and that it follows from an inversion that abundant machine checking produces. We distinguish three layers of verification: derivational validity, which a kernel checks; representational fidelity, whether the formal statement means the informal question; and epistemic significance. Only the first is mechanizable, and we give the reason, inherited from Fetzer, why the second is not mechanizable even in principle. Driving the first to zero cost does not reduce the verification burden on a body of knowledge; it transfers that burden onto layers whose throughput is fixed by the supply of qualified readers.

We support this with measurements of the published artefacts. In the August corpus the kernel checked proofs total 20.6 MB while the statements requiring human audit total 55.6 KB, a ratio of 379 to 1, and those statements introduce 218 bespoke definitions rather than relying on the community vetted library. The audit surface is therefore small but irreducibly expert. We conclude that engineering can shrink the audit surface and cannot manufacture the auditor, and that the resulting scarcity is not of verification but of adjudication. We offer a six category taxonomy of representational mismatch usable as an audit protocol, a disclosure schema for machine generated discovery claims, and an analysis of what transfers to software, cryptography and regulated decision systems.
\end{abstract}



\section{Two results, two fates}

On 20 May 2026 OpenAI announced that a general purpose reasoning model had disproved the Erdős unit distance conjecture, a problem posed in 1946 concerning the maximum number of pairs of points at distance exactly one among $n$ points in the plane (OpenAI 2026a). The prevailing expectation was that rescaled grid constructions were essentially optimal and that the answer was $n^{1+o(1)}$. The model produced an infinite family of configurations exceeding that bound by a polynomial factor.

What happened next is the part that matters here. On the same day, Noga Alon, Thomas Bloom, W.\ T.\ Gowers, Daniel Litt and Will Sawin posted a paper whose abstract begins: ``We present a short, digested, human-verified version of the recent OpenAI-generated counterexample to the Erdős unit distance conjecture, and a sequence of reflections on it'' (Alon et al.\ 2026). They situated the argument in existing mathematics, attributing its crucial ideas to Ellenberg-Venkatesh, Golod-Shafarevich, and Hajir-Maire-Ramakrishna. Sawin subsequently supplied an explicit constant. Gowers indicated he would recommend the result for the \emph{Annals of Mathematics}. The claim entered the literature by the ordinary route, and the operative verb in the abstract is \emph{digested}: five mathematicians converted a machine's output into a form humans could evaluate, and then evaluated it.

On 1 August 2026 the same laboratory published \emph{Ten Advances in Mathematics and Theoretical Computer Science}, reporting results from an internal model on ten problems, most open for decades (OpenAI 2026b). The presentation was in one respect far stronger than in May. Every result shipped with a Lean 4 formalization, released under Apache 2.0 as the repository \texttt{openai/ten-proofs}. The project metadata records no unproved steps anywhere in the development and permits only the three standard axioms of the mathematical library, \texttt{propext}, \texttt{Classical.choice} and \texttt{Quot.sound} (OpenAI 2026c). The repository ships configurations for Comparator, an independent checker, with a second independent kernel enabled. By the standard the formal methods community has advocated for forty years, this is close to exemplary.

Four weeks later, the fourth of those results, a claimed counterexample to Connes's rigidity conjecture, sits in an unresolved dispute. A critique appeared within a day claiming to have traced the entire development and to have found that the constructed groups do not satisfy the infinite conjugacy class condition the conjecture requires. The critique was itself contested: a mathematician at the laboratory rebutted it publicly, commentators identified an apparent error in its group theory, and the document's provenance came under question, its stated institution appearing to exist only as a website and the text itself appearing machine generated. A separate audit of the critique's revision history has since been posted. At least three mutually independent machine generated counterexamples to the same conjecture are now in circulation from different laboratories. Some participants in the public discussion resolved the question, to their own satisfaction, by asking language models which side was right.

We take no position on the mathematics. Whether the August construction is correct is not ours to judge and, we will argue, not necessary to judge. The object of interest is the difference in fate. Two claims, one laboratory, one year, comparable stakes; in one case a rapid and conclusive community verdict, in the other a dispute that machine checkable certificates did not settle and may have helped to prolong.

The natural reading is that the May result was easier, or luckier, or that the August results are simply harder to check because there is more of them. We will show that this reading is wrong in an instructive way. Measured directly, the August artefacts present a \emph{smaller} human reading task per result than the May argument did. The difficulty is not volume. It is that the reading which remains can be done only by a very small number of people, and in August those people were neither recruited nor, as the laboratory's own metadata records, replaced by anything other than another model.

Our thesis is that this is the general case. Making derivational checking free does not reduce the verification burden attached to a body of knowledge. It transfers that burden onto layers that cannot be mechanized and whose throughput is bounded by the supply of qualified readers, while removing the only limit that previously constrained the rate at which claims arrived. The result is not faster knowledge but a widening gap between what has been certified and what has been adjudicated.

Section 2 sets out the three layer distinction the argument needs. Section 3 recovers the reason, from a debate that ran in the \emph{Communications of the ACM} between 1979 and 1988, why the middle layer resists mechanization in principle. Section 4 offers a taxonomy of the ways formal statements come apart from informal questions, grounded in documented incidents. Section 5 reports measurements of the August corpus. Section 6 states the inversion. Section 7 examines the strongest engineering response now available and what it does not reach. Sections 8 and 9 consider transfer and remedies.


\section{Three layers of verification}

It is useful to separate three questions that the phrase ``the proof has been verified'' runs together.

\textbf{L1, derivational validity.} Does the proof term type check against the kernel? Every inference is checked mechanically against a small trusted core. This is what a proof assistant does, and it does it very well. The cost of performing an L1 check, given the artefact, tends towards zero and is falling. The August corpus is exemplary at this layer, and we will insist on the point: zero unproved steps, three standard axioms, independently rechecked by a second kernel.

\textbf{L2, representational fidelity.} Does the formal statement, together with every definition it rests on, mean the informal question? This asks whether the sentence the kernel certified is the sentence the field cares about. No compiler checks it. It is checked, if at all, by a human who knows both the mathematics and the formal language, reading the statement and judging.

\textbf{L3, epistemic significance.} Is the result novel, does it matter, is it framed correctly, does it advance the subfield? This is the traditional business of referees and of the discipline over time, and it is not our main subject, though it shares L2's dependence on scarce human attention.

The three layers differ in three respects that the rest of the paper turns on: in what checks them, a machine, an expert, a community; in whether they can in principle be mechanized, yes, no, no; and in how their cost behaves as the volume of claims rises, falling to zero, roughly constant per artefact, and rising.

Two clarifications forestall predictable objections. First, this is not a claim that proof assistants are untrustworthy. The opposite: the L1 guarantee is the strongest guarantee in the epistemology of mathematics, and the argument here depends on it being strong. Second, the distinction is not new as an observation. Practitioners have always known it, it is discussed carefully in the formalization literature (Avigad 2023; Commelin and Topaz 2023; Bayer et al.\ 2022), and since August 2026 it has become common currency in public commentary. What is new here is the treatment of the three layers as an economic structure with different scaling behaviour, and the consequences that follow.

The autoformalization programme is the natural place to look for a mechanical solution to L2, and it is worth saying at the outset why it does not supply one. That programme aims to translate informal mathematics into formal statements automatically (Szegedy 2020; Wu et al.\ 2022), and is evaluated against benchmark suites of paired informal and formal statements (Zheng et al.\ 2021; Azerbayev et al.\ 2023). The evaluation is where the difficulty sits. A benchmark of paired statements encodes, in its pairings, precisely the correspondence whose reliability is in question, and those pairings were fixed by human judgement when the benchmark was built. Autoformalization can therefore produce candidate statements at scale, and can be scored against a fixed stock of human judgements, but it cannot certify a correspondence for a question nobody has yet paired. The regress set out in Section 3 applies to it directly.


\section{What the 1979 to 1988 debate already settled}

The problem now being rediscovered was posed and largely resolved, at the level of principle, in an argument that ran through the \emph{Communications of the ACM} across a decade.

De Millo, Lipton and Perlis (1979) argued that the certainty of mathematics is produced by a social process. A theorem becomes reliable because it is read, doubted, simplified, taught, generalized and used, by many people over time. Formal program verification, they argued, cannot inherit that reliability, because the verifications are long, mechanical and unreadable, and so are never subjected to the process that confers belief. The argument was widely read as hostile to formal methods and has often been treated as refuted by the subsequent success of proof assistants. That reading misses what survives. Their central claim was not that mechanical checking fails; it was that mechanical checking is not the thing that makes mathematics trustworthy, because trustworthiness is a property conferred by readers, and unreadable objects have none.

Fetzer (1988) sharpened the point into a distinction that our L1 and L2 track directly. Formal verification establishes relations among formal objects. It cannot establish the correspondence between a formal object and the extra formal thing it is meant to represent, because that correspondence is not itself a formal relation and so is not the kind of thing a derivation can bear on. In Fetzer's case the extra formal thing was the behaviour of a physical machine. In ours it is the content of a mathematical question as the discipline understands it, which is transmitted through papers, seminars, examples and usage rather than through any canonical formal text.

This yields what we will call the \textbf{grounding regress}. Suppose we wish to check mechanically that a formal statement $S$ faithfully renders an informal question $Q$. A mechanical check requires both relata to be formal. So it requires a formal rendering of $Q$, call it $S'$. But then the question of whether $S'$ faithfully renders $Q$ is exactly the question we began with. Either the regress continues or it terminates in an unaided human judgement that some formal object means some informal thing. It always terminates in the latter, because that is the only place it can terminate.

The consequence is stronger than the familiar observation that L2 is hard, or neglected, or not yet automated. L2 is not a task awaiting a better tool. It is the point at which formal method necessarily touches something outside itself, and machine checking's guarantee, however strong, stops exactly there. Every improvement in automation makes the terminal human judgement more consequential, because more weight rests on it and less independent evidence surrounds it.

MacKenzie (2001) documented how this played out sociologically across the first decades of mechanized proof, and Arkoudas and Bringsjord (2007) examined the epistemology of computer assisted mathematical justification directly. Our contribution in this section is not the argument, which is Fetzer's, but its transposition: what was a claim about programs and physical machines becomes, under machine generated mathematics, a claim about conjectures and disciplinary meaning, and it becomes quantitative for the first time, because for the first time the number of certified claims is not bounded by the number of people willing to produce them.


\section{A taxonomy of representational mismatch}

If L2 must be checked by hand, it is worth knowing what to look for. The following taxonomy is derived from documented incidents in the formalization record and from structural features of the August corpus. We present it as an audit protocol rather than as a catalogue of defects: each category names an obligation an auditor must discharge, not an accusation against any particular development.

\textbf{T1, quantifier scope displacement.} The formal statement binds a variable at a different scope than the informal claim requires. The clearest documented instance comes from the Flyspeck project, the formal proof of the Kepler conjecture and the most carefully executed formalization effort of its generation. Scharf (2017) observed that Hales establishes that the density of a packing within a ball of radius $r$ is bounded by $\pi/\sqrt{18}+c/r$, where the constant $c$ depends on the packing, whereas the intended statement requires a single constant independent of the packing. The difference is the order of two quantifiers. Nothing about the mechanical check reveals it, and the project in which it occurred was neither rushed nor careless.

\textbf{T2, reformulation of the target.} The formal statement is a deliberately chosen surrogate for the informal one, adopted for tractability, with the equivalence argued informally or not at all. Flyspeck again supplies the example, and Hales is explicit about it: rather than formalize density as a limit, the project formalized a statement about density within a finite container with an error term whose ratio to the container's volume vanishes, recovering the traditional statement in the limit. This is legitimate and was declared. The audit obligation is that someone must check the recovery, and that check is informal. Hales (2024) revisits the project critically at a decade's distance, and the retrospective is worth reading as evidence of how much of the difficulty in a formalization lies outside what the kernel sees.

\textbf{T3, bespoke definitional substitution.} The statement defines its own version of a standard notion rather than importing the community vetted one. The imported definition carries the weight of many eyes; a local definition carries none, and shifts the auditor's task from reading a theorem to reading a theorem plus a dictionary. Section 5 shows that the August corpus does this systematically, introducing 218 local definitions across twelve statement files.

\textbf{T4, domain or universe restriction.} The formal statement quantifies over a narrower class than the informal claim. This is easy to introduce accidentally in dependently typed systems, where universe levels must be chosen. To take an observable example from the August corpus, the Connes rigidity statement defines Kazhdan's property (T) by quantifying over unitary representations on Hilbert spaces in the same universe as the group. For a group in the lowest universe this restricts the representations considered. Whether that restriction is harmless for the intended claim is precisely the sort of question the kernel does not raise and an expert must settle. We report it as an audit obligation, not a defect.

\textbf{T5, hypothesis satisfied by construction rather than by content.} The statement carries the right hypothesis, and the development proves the constructed object satisfies it, but the proof establishes satisfaction of the formal predicate under a construction whose relation to the intended objects is what is in doubt. This is the shape of the contested Connes case, and it is worth stating precisely because the public account has been widely garbled. The formal statement does contain the infinite conjugacy class condition as an explicit conjunct, and defines it standardly, as infinitude of the group together with infinitude of every non-identity conjugacy class. The dispute is not about a missing hypothesis. It concerns whether the lemmas discharging that condition, argued for one class of group extension, apply to the construction actually used. An auditor cannot discharge T5 by reading the statement alone, which is what makes it the most expensive category in the taxonomy.

\textbf{T6, encoding substitution.} For claims about computation, the statement rests on a bespoke encoding of machines, languages or complexity classes, and is faithful only to the extent the encoding is. The August closest vector problem statement defines its own bit level Turing machines, verifiers and membership in NP, forty local definitions in all. A hardness claim is exactly as meaningful as the definition of hardness it uses, and here that definition is local.

T1 and T2 are attested in the historical record. T3, T4 and T6 are observable properties of the August corpus, reported below. T5 is the category the current dispute occupies. We do not claim the taxonomy is exhaustive; we claim it is sufficient to structure an audit, and that no comparable taxonomy currently exists. The nearest adjacent work, Mao et al.\ (2026), concerns whether a formal \emph{proof} mirrors the reasoning of a natural language argument, a different and complementary question, and proposes no taxonomy of statement level mismatch.


\section{The audit surface: measurements}

The intuition that machine generated proofs overwhelm human checkers by sheer size is widespread and, we will show, wrong in a way that matters. We measured the published corpus directly. All figures are from \texttt{openai/ten-proofs} at main, repository last updated 5 August 2026, retrieved and archived 28 August 2026; they are reproducible from the repository's file tree.

The repository separates two kinds of file. Ten proof modules carry the developments themselves. Twelve challenge files, one per headline declaration, carry the statements in isolation, intended for independent checking. The separation is deliberate and, as we argue in Section 7, admirable.

The proof corpus totals 21,581,745 bytes, roughly 20.6 MB. The statement corpus totals 56,901 bytes, roughly 55.6 KB. The ratio is \textbf{379 to 1}: the material a human must read to discharge L2 is \textbf{0.263 per cent} of the released Lean. Per result the statements run from 1,117 bytes for the non-sofic group construction to 10,839 bytes for the closest vector problem. The contested Connes statement is 7,483 bytes, 223 lines.

So the audit surface is not large. Two hundred and twenty three lines is an afternoon's work for the right reader. The engineering did what it was designed to do, reducing the human reading task by more than two orders of magnitude.

The second measurement explains why this did not suffice. Across the twelve statement files there are \textbf{218 local definitions} supporting 38 theorem statements. Every file imports the whole mathematical library, and then declines to use its definitions of the notions at issue. The spherical codes statement introduces 60 definitions for 4 theorems; the closest vector statement 40 for 4, including bespoke Turing machines, verifiers and a local definition of NP; the Connes statement 22, including local definitions of the infinite conjugacy class condition, property (T), the group von Neumann algebra, the canonical trace and trace preserving isomorphism. At the other end, the multicolour Ramsey statement needs only 3 and the non-sofic group statement only 3.

This changes what the audit surface consists of. The reader is not asked to check that a theorem correctly composes notions the community has already vetted. The reader is asked to check a theorem \emph{and} to validate a private dictionary of 218 terms, each of which must be compared against its standard meaning. That is T3 at scale, and it converts a short document into an expert task. Reading 223 lines of Lean is quick. Certifying that this particular rendering of property (T), this particular construction of the group von Neumann algebra and this particular notion of trace preserving isomorphism jointly reconstruct what Connes conjectured requires a specialist in von Neumann algebras who also reads Lean fluently and is willing to spend days adjudicating another organization's claim. That intersection is very small, and it does not grow when generation capacity grows.

The third observation is documentary rather than quantitative, and it is the most direct evidence available. The repository's own metadata file records the review status of the entire development as agent-reviewed. The laboratory's records state that the layer we have called L2 was not checked by a human domain expert.

Set beside the May case, the contrast is exact. There, five named mathematicians produced a digested, human verified version and placed the argument in its literature, within a day. Here, twelve declarations were released with a machine in the reviewer's chair. The difference in outcome followed.


\section{The inversion}

We can now state the argument compactly.

Let $g$ denote the cost of generating one certified claim, $a$ the cost of a competent L2 audit of one claim, and $H$ the expert reading hours available per period to a given subfield. Certified claims can be produced at a rate bounded by budget divided by $g$. They can be audited at a rate bounded by $H/a$.

Machine generation drives $g$ towards zero, and the August corpus reports a per problem inference cost below two thousand dollars at internal rates. Nothing drives $a$ towards zero. The measurements above show why: engineering can compress the artefact, as the challenge file separation did by a factor of 379, but the residue is irreducibly expert, and expertise is what $a$ is made of. Nor does $H$ grow. The population able to audit a claim in von Neumann algebras is fixed on any timescale relevant here, is not compensated for the work, and is drawn from the same people who would otherwise be proving theorems.

Two consequences follow.

The first is the \textbf{transfer of burden}. Total verification work per claim does not fall when L1 becomes free; its composition changes. Before, a reader who checked a proof line by line thereby also checked, in passing, that the statement said something sensible, because reading the argument required understanding the objects. Mechanical checking removes the line by line reading and with it the incidental L2 check that came free with it. The L2 obligation is not reduced by automation; it is \emph{isolated} by automation, and left standing alone.

The second is the \textbf{decoupling of certification from belief}. As the ratio of produced claims to audited claims diverges, the fact that a claim is certified stops carrying information about whether anyone has judged it to mean what it says. Certification remains a perfectly reliable signal about L1 and becomes an increasingly uninformative signal about the thing a reader actually wants to know.

The observable consequence is what we call \textbf{adjudication scarcity}. When a dispute arises about a certified claim, resolving it requires exactly the resource that has become the bottleneck. If that resource does not appear, the dispute does not resolve; it degrades. The August episode shows the degradation in an unusually complete form. Within four weeks the community had produced a critique of the claim, a rebuttal of the critique, an investigation into the provenance and revision history of the critique's author, an argument about the existence of that author's institution, two further machine generated counterexamples to the same conjecture from other laboratories, and at least some participants adjudicating by asking language models. What it had not produced, so far as the public record shows, is one qualified specialist reading the 223 line statement and saying what it means.

This is the sense in which verification abundance produces adjudication scarcity. The claims arrive certified, which makes them hard to dismiss; there are many of them, which makes them hard to prioritize; the qualified readers are unchanged in number and now face a task from which the incidental rewards of reading a proof, learning the technique, have been stripped out. The community's fallback, when adjudication fails, is testimony, and the available testimony increasingly comes from systems of the same kind that generated the claim.

We note explicitly that this argument does not depend on the outcome of the Connes dispute. If the August construction is correct, the community took more than four weeks and did not establish that it was correct. If it is flawed, the community took more than four weeks and did not establish that it was flawed. Either way the adjudication failed, and it is the failure, not the mathematics, that we are describing.


\section{What good engineering reaches, and what it does not}

It would be easy and wrong to present the August release as careless. It is, by the standards of the field, unusually conscientious, and the paper's argument is stronger for saying so.

Four features deserve credit. The development contains no unproved steps and depends on only the three standard axioms, so the L1 guarantee is as strong as the technology permits. The statements are separated from the proofs into small files, which is precisely the right response to the reading problem and reduces the audit surface by a factor of 379. Independent checking is supported through configurations for Comparator, with a second independent kernel enabled, so the trusted base does not reduce to one implementation. And the metadata is candid: sorry counts, axiom lists, wall clock time, prior work credited to two existing community formalization projects, and the review status recorded as agent-reviewed rather than obscured.

That last point cuts both ways, and honestly. The laboratory disclosed the thing that most damages the release. Our argument uses its own record against its own claim, which is only possible because the record was published.

What the engineering does not reach is the terminal judgement. Comparator can confirm that the proof establishes the challenge statement, using an independent kernel. It cannot confirm that the challenge statement is Connes's conjecture. The separation of statement from proof relocates the human reading and makes it tractable; it does not perform it. This is not a deficiency of the tool. By the argument of Section 3 no tool can perform it, and a tool that claimed to would be making a category error.

The measurements sharpen the point. Having compressed the audit surface by two orders of magnitude, the release then expanded it again along a different axis by introducing 218 local definitions in place of library ones. The compression was in bytes; the expansion was in expertise. Since expertise, not bytes, is what bounds $a$, the net effect on auditability was small. A development that used library definitions wherever they existed would have a larger byte count in its statement files and a substantially smaller audit surface in the sense that matters. That is an actionable recommendation, and it follows from measurement rather than from taste.


\section{Transfer}

Nothing in the argument is specific to mathematics. It requires only that a generator produce artefacts, that a mechanical checker validate them against a specification, and that the correspondence between specification and intent be settled by a human. Those conditions are met wherever formal methods meet machine generation.

In \textbf{software and cryptography} the correspondence is the specification, and the transfer is exact: a machine checked proof that an implementation meets a specification says nothing about whether the specification captures the security property intended. As machine generated code grows and formal pipelines mature, the specification review becomes the binding constraint, and specification review is scarce, expert and unrewarded in exactly the way L2 audit is. The August corpus contains a hardness result for the closest vector problem, which underlies lattice based post quantum cryptography, and states it in terms of a locally defined notion of NP hardness. Whatever the merits of that development, the governance lesson is that a cryptographic hardness claim should be routed to specification review, not accepted on the strength of a certificate.

In \textbf{regulated decision systems}, compliance logic, financial models, actuarial and audit rules, the same structure appears with a weaker mechanical layer and a stronger institutional one. Machine checkable representations of correctness are partial at best, so most of the burden already sits at L2 and L3. Here the August episode is a caution rather than a template: the lesson to carry across is not that formal verification gates should be installed everywhere, but that a mechanical pass must never be reported as though it settled a question of meaning.

The general recommendation is architectural and modest. Treat the generator as a hypothesis producer. Keep the checker independent of it, which the Comparator arrangement does well. And staff the specification review explicitly, as a named role with named people and allocated time, because the analysis above says it is the layer that will silently fail to happen.


\section{Remedies}

We suggest three, in increasing order of ambition.

\textbf{Disclosure.} The August release reports a per problem inference cost and no denominator: how many problems were attempted, how many failed, how they were selected. Without a denominator no external party can compute a success rate, and without a success rate no institution can estimate how much audit capacity a given generation capacity will demand. We propose that machine generated discovery claims report, at minimum: problems attempted and abandoned; the selection procedure; total inference spend including failures; human hours spent on formalization and on statement review; who performed the statement review and with what qualification; and which definitions are local rather than imported from a vetted library. Most of these the August release already tracks internally; the last is mechanically extractable, as Section 5 demonstrates.

\textbf{A named role.} The analysis identifies a job that currently belongs to nobody: reading the formal statement and certifying that it renders the intended question. It requires domain expertise and formal fluency, it is distinct from refereeing a proof, and it is the only step in the pipeline that cannot be automated. It should be named, credited and, where institutions can manage it, paid. Journals accepting formalized results should require it and record who did it, exactly as the \texttt{formalization.yaml} schema already records who checked what.

\textbf{Assurance levels.} Because L2 admits degrees of scrutiny rather than a binary, claims could carry a level: kernel checked only; kernel checked with statements separated for independent audit; statement audited by a named domain expert; statement audited and the informal to formal correspondence argued in print. The August release sits at the second level and reports as much. The May result, in effect, reached the fourth, by a route that predates the vocabulary: five mathematicians digested the argument into human form and published the digestion.


\section{Limitations}

The empirical base is two cases from one laboratory in one year, chosen because they are unusually well documented and closely matched, not because they are representative. The measurements in Section 5 are of published artefacts and say nothing about unpublished internal review; the agent-reviewed metadata is evidence about what was recorded, not proof that no human ever read the statements. Byte counts are a crude proxy for reading effort, and the local definition count, while mechanically extracted and reproducible, does not weight definitions by difficulty; a fuller treatment would count tokens and grade definitions by depth. The Connes dispute is live and may resolve in either direction after this is written, which is why the argument has been constructed not to depend on its resolution. The taxonomy of Section 4 is grounded in documented incidents but has not been validated by applying it systematically to a large corpus, which is the obvious next study. Finally, the claim that expert audit capacity is fixed is an assumption about a timescale of years; over decades formal fluency among mathematicians is plainly rising, and the argument is about the transition rather than the steady state.


\section{Conclusion}

The formal methods programme has spent forty years arguing that mechanical checking should replace human trust in long arguments, and it has largely won. The August 2026 release is what winning looks like: ten results, no unproved steps, three axioms, an independent kernel, statements separated for audit, metadata published. The L1 problem is solved.

The consequence is not the one the programme anticipated. Solving L1 did not reduce the amount of human judgement mathematics requires. It isolated that judgement, stripped away the incidental checking that used to come free with reading a proof, and removed the constraint that previously limited how fast claims could arrive. What remains is a small, irreducibly expert reading task, 0.263 per cent of the artefact by size and nearly all of it by difficulty, which in the case examined here nobody performed.

De Millo, Lipton and Perlis were right that unread proofs confer no belief, and wrong to conclude that mechanization was therefore futile; mechanization works, and the objects it produces are still unread. Fetzer was right that verification cannot reach across the boundary between formal and intended, and the boundary has now been reached at industrial volume. What is new is only the arithmetic. As long as producing certified claims required people, the supply of claims and the supply of readers were drawn from the same pool and scaled together. That coupling is now broken, and nothing has replaced it.

The scarce resource in mathematical knowledge production is no longer proof. It is adjudication.


\end{document}